\RequirePackage{lineno}
\documentclass[10pt,journal,compsoc,onecolumn]{IEEEtran}
\ifCLASSOPTIONcompsoc
  \usepackage[nocompress]{cite}
\else
  \usepackage{cite}
\fi
\ifCLASSINFOpdf
\else
\fi

\usepackage{array}
\usepackage{amsmath}
\usepackage{mathrsfs}
\usepackage[utf8]{inputenc} % allow utf-8 input
\usepackage[T1]{fontenc}    % use 8-bit T1 fonts
\usepackage{hyperref}       % hyperlinks
\usepackage{url}            % simple URL typesetting
\usepackage{booktabs}       % professional-quality tables
\usepackage{amsfonts}       % blackboard math symbols
\usepackage{nicefrac}       % compact symbols for 1/2, etc.
\usepackage{microtype}      % microtypography
\usepackage{lipsum}
\usepackage{graphicx}
\graphicspath{ {./images/} }
\usepackage{blindtext}
\usepackage[inline]{enumitem}
\usepackage[dvipsnames]{xcolor}
\usepackage{float}
\begin{document}
%\setpagewiselinenumbers

\title{StressNet: Deep Learning to Predict Stress With Fracture Propagation in Brittle Materials}

\author{Yinan Wang,~\IEEEmembership{}
        Diane Oyen,~\IEEEmembership{}
        Weihong (Grace) Guo,~\IEEEmembership{}
        Anishi Mehta,~\IEEEmembership{}
        Cory Braker Scott,~\IEEEmembership{}
        Nishant Panda,~\IEEEmembership{}
        \\M. Giselle Fernández-Godino,~\IEEEmembership{}
        Gowri Srinivasan,~\IEEEmembership{}
        and~Xiaowei Yue~\IEEEmembership{}% <-this % stops a space
        
\IEEEcompsocitemizethanks{

\IEEEcompsocthanksitem Yinan Wang and Xiaowei Yue are with the Grado
Department of Industrial and Systems Engineering, Virginia Tech, Blacksburg, VA, 24061.\protect\\
E-mail: \{yinanw, xwy\}@vt.edu

\IEEEcompsocthanksitem Diane Oyen, Nishant Panda, M. Giselle Fernández-Godino and Gowri Srinivasan are with Los Alamos National Laboratory, Los Alamos, NM, 87544\protect\\
E-mail: \{doyen, nishpan, gisellefernandez, gowri\}@lanl.gov

\IEEEcompsocthanksitem Weihong (Grace) Guo is with the Department of Industrial and Systems Engineering, Rutgers University, New Brunswick, NJ, 08901\protect\\
E-mail: wg152@soe.rutgers.edu

\IEEEcompsocthanksitem Anishi Mehta is with the College of Computing, Georgia Institute of Technology, GA, 30332 \protect\\
E-mail: anishi.mehta@gatech.edu

\IEEEcompsocthanksitem Cory Braker Scott is with the Department of Computer Science, University of California Irvine, Irvine, CA, 92697 \protect\\
E-mail: scottcb@uci.edu
}% <-this % stops an unwanted space
\thanks{Manuscript received xxxx; revised xxxx.}
\thanks{(Corresponding author: Xiaowei Yue)}}


\markboth{Manuscript}%
{Shell \MakeLowercase{\textit{et al.}}: Bare Demo of IEEEtran.cls for Computer Society Journals}




\IEEEtitleabstractindextext{%

\begin{abstract}
Catastrophic failure in brittle materials is often due to the rapid growth and coalescence of cracks aided by high internal stresses. Hence, accurate prediction of maximum internal stress is critical to predicting time to failure and improving the fracture resistance and reliability of materials. Existing high-fidelity methods, such as the Finite-Discrete Element Model (FDEM), are limited by their high computational cost. Therefore, to reduce computational cost while preserving accuracy, a novel deep learning model, "StressNet," is proposed to predict the entire sequence of maximum internal stress based on fracture propagation and the initial stress data. More specifically, the Temporal Independent Convolutional Neural Network (TI-CNN) is designed to capture the spatial features of fractures like fracture path and spall regions, and the Bidirectional Long Short-term Memory (Bi-LSTM) Network is adapted to capture the temporal features.  By fusing these features, the evolution in time of the maximum internal stress can be accurately predicted. Moreover, an adaptive loss function is designed by dynamically integrating the Mean Squared Error (MSE) and the Mean Absolute Percentage Error (MAPE), to reflect the  fluctuations in maximum internal stress. After training, the proposed model is able to compute accurate multi-step predictions of maximum internal stress in approximately 20 seconds, as compared to the FDEM run time of $4$ hours,  with an average MAPE of  $2\%$ relative to test data.

\end{abstract}

}

\maketitle

\IEEEdisplaynontitleabstractindextext

% \fi
\IEEEpeerreviewmaketitle

\IEEEraisesectionheading{\section{Introduction}\label{sec:introduction}}
%\linenumbers
\IEEEPARstart{B}{rittle} materials, such as glass, ceramics, concrete,  \textcolor{Black}{some metals}, and composite materials, are widely used in many applications \textcolor{Black}{that involve} complex dynamics, impulse, or shock loadings. \textcolor{Black}{In structural materials, }high-stress concentrat\textcolor{Black}{ion} around \textcolor{Black}{micro-scale} defects precipitates cracks, eventually \textcolor{Black}{leading to fracture initiation, propagation, and coalescence}. \textcolor{Black}{In brittle materials,} fractures propagate fast \textcolor{Black}{with almost no elastic deformation leading to catastrophic failure with little warning. The dynamics of fracture evolution are governed strongly by maximum internal stresses in the material}. However, accurate prediction of \textcolor{Black}{maximum} internal stress of brittle material under dynamic \textcolor{Black}{loading conditions} remains a challenge in the field of material science \cite{doi:10.1098/rsta.2016.0436, Zhou2018, Perez2004}. \textcolor{Black}{Therefore, e}nsuring the durability and reliability of brittle materials under various dynamic loading \textcolor{Black}{conditions is imperative, especially in cases where accidents can jeopardize safety and security}.

A material fails when the maximum internal stress in any direction \textcolor{Black}{equals} either the tensile or compressive strength\cite{Leckie:2044410}. Under idealized conditions, the internal stress field should be distributed \textcolor{Black}{homogeneously} through the sample. However, real-world materials inevitably contain microfractures, \textcolor{Black}{defects, or impurities, which result in high values of stresses being concentrated internally} \cite{doi:10.1111/ffe.12228, Durelli1962, SAPORA2015296}. Hence, the real fracture strength of a \textcolor{Black}{brittle} material is always lower than the theoretical value. \textcolor{Black}{The presence of cracks increases stress values locally, and in turn, the stress concentration around fractures results in the fractures propagation. Predicting the maximum internal stress of \textcolor{Black}{a} material is extremely difficult because the stress and damage are highly coupled.}

\begin{figure}[ht]
    \includegraphics[width=0.55\columnwidth]{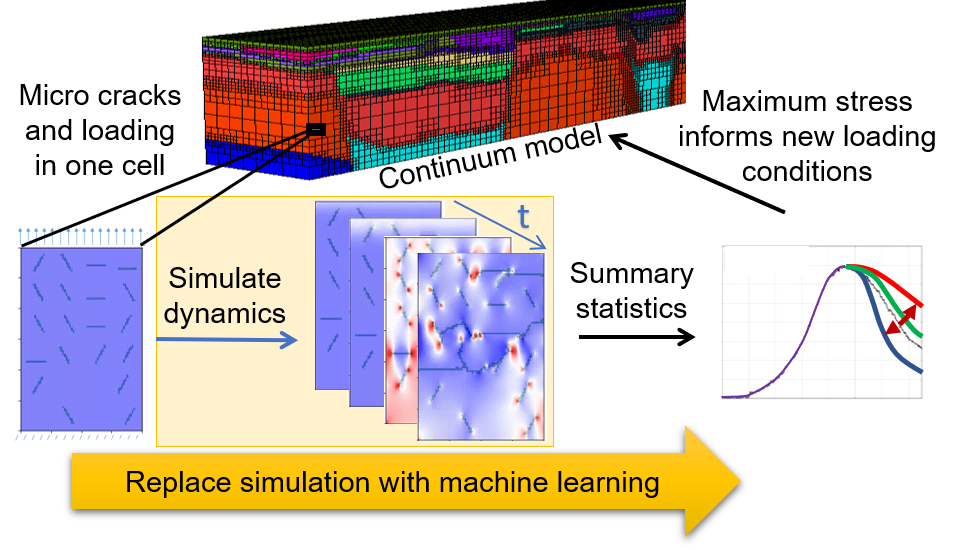}
    \centering 
    \caption{\textcolor{Black}{Proposed Workflow. The machine learning \textcolor{Black}{model} informs the continuum model with crack statistics and stresses.} The machine learning model replace\textcolor{Black}{s} the \textcolor{Black}{expensive mesoscale} simulation model (\textcolor{Black}{high-fidelity) to speed up predictions while maintaining its accuracy.}}
    \label{system_overview}
\end{figure}

A common approach to simulate the stress and strain of a given \textcolor{Black}{material} is the finite element method (FEM) \cite{Ashcroft2011,7058481}. The main idea of FEM lies in simplifying the problem by breaking the material down into a large number of finite elements and then building up an algebraic equation to compute the coupled mechanical deformations and stresses based on the boundary and load conditions. The Hybrid Optimization Software Suite (HOSS)\textcolor{Black}{, developed at Los Alamos National Laboratory (LANL),} is a hybrid finite-discrete element method (FDEM) that can simulate the fracture growth of \textcolor{Black}{both 2D and 3D} physical systems \cite{SCHWARZER2019322, osti_1369045}. \textcolor{Black}{Within HOSS simulations, }the material is modeled  \textcolor{Black}{as finite elements and the fractures are represented by discrete elements which can only form along the boundaries of the finite elements}. Although this method gives accurate predictions of the fracture growth and the dynamics of stress distribution, \textcolor{Black}{it} is \textcolor{Black}{computationally} intensive, \textcolor{Black}{especially when multiple runs are needed to obtain the statistical variability naturally existent in real-world materials}. \textcolor{Black}{Machine learning (ML) techniques are becoming popular \cite{Srinivasan2018} in modeling complex systems because they can serve as lower-order surrogates to approximate higher-fidelity models, which significantly reduces the model complexity and computation time, as shown in Fig.~\ref{system_overview}}.

\textcolor{Black}{Despite recent advances, a}pplying ML models for predicting the maximum \textcolor{Black}{internal} stress with fracture \textcolor{Black}{propagation} of materials is still limited. \textcolor{Black}{Nash et al. \cite{Nash2018} reviewed the most recent deep learning methods for detection, modeling, and planning for material deterioration.} Nie \textcolor{Black}{et al.} \cite{DBLP:journals/corr/abs-1808-08914} used Encoder-Decoder Structure based on Convolutional Neural Network (CNN) to generate the stress field in cantilevered structures. However, \textcolor{Black}{these methods do not consider} temporal dynamics of the stress field or fracture within the material. On the other hand, most recent papers focus on predicting the \textcolor{Black}{fracture propagation} instead of internal stresses. Rovinelli et al. \cite{Rovinelli2018} built a Bayesian Network (BN) to identify an analytical relationship between crack propagation and its driving force, which focuses on predicting the direction of crack propagation instead of the detailed crack path. Hunter et al. \cite{HUNTER201987} applied an Artificial Neural Network (ANN) to \textcolor{Black}{approximately} learn the dominant trends and effects that can determine the overall material response. Moore et al. \cite{MOORE201846} implemented a Random Forest (RF) and a Decision Tree (DT) to predict the dominant fracture path within the material. Shi  \cite{SHI2008588} compared the performance of Support Vector Machine (SVM) and ANN in fracture prediction. Schwarzer et al. \cite{SCHWARZER2019322} employed a \textcolor{Black}{Graph Convolutional Network to recognize} features of \textcolor{Black}{the fractured} material and a \textcolor{Black}{recurrent neural network (RNN)} \textcolor{Black}{to model} the evolution of these features. \textcolor{Black}{Fern\'andez-Godino et al. \cite{fernandez2020accelerating}} use \textcolor{Black}{\textcolor{Black}{a} RNN} to bridge meso and continuum scales for accelerating predictions in a high strain rate application problem. \textcolor{Black}{} One common issue \textcolor{Black}{with previous \textcolor{Black}{works}}  (\cite{HUNTER201987,MOORE201846,SHI2008588,SCHWARZER2019322,fernandez2020accelerating}) is that the models are built on manually selected features such as fracture length, orientation, distance between fractures, etc., instead of features learned from the raw data. Manually \textcolor{Black}{selected} features could reduce the computation requirement, \textcolor{Black}{but }it might cause information \textcolor{Black}{loss} and degrade model performance.

\textcolor{Black}{The proposed work seeks to go beyond existing methods by considering a dynamically evolving stress tensor. Simulation results from HOSS, which provide the data for building and validating ML surrogates, have} two major properties. \textcolor{Black}{First, at each time step,} a 3-way tensor representing the \textcolor{Black}{spatial properties of fractures} and the stress field. \textcolor{Black}{Additionally, the entire simulation is }a time-series representing the temporal dependencies among different time-steps. \textcolor{Black}{The relevant literature about extracting spatial features of the tensor data and temporal dependencies of time series data is further reviewed.} 

For the tensor data, Yue et al. \cite{doi:10.1080/00401706.2019.1592783} proposed a tensor mixed-effects (TME) model to analyze massive high-dimensional Raman mapping data with a complex correlation structure. Yan et al. \cite{DBLP:journals/corr/abs-1801-07455} and Si et al. \cite{DBLP:journals/corr/abs-1805-02335} applied Graph Convolutional Neural Networks (GCN) to capture the spatial structure information of a body skeleton to recognize different actions in \textcolor{Black}{the }video. Shou et al. \cite{DBLP:journals/corr/ShouCZMC17} implemented a 3D Convolutional-De-Convolutional (CDC) structure to \textcolor{Black}{detect and }localize actions in \textcolor{Black}{the }video. All of the aforementioned work proposed novel \textcolor{Black}{methods} to extract features from tensor data for different tasks. \textcolor{Black}{However, these methods do not consider temporal evolution. Therefore, they cannot be used directly to predict maximum internal stresses with fracture propagation.}

For the analysis of time series data, in the context of statistical learning, Auto-Regressive Integrated Moving Average (ARIMA) is a class of models that captures a suite of different standard temporal structures in time series data \cite{10.2307/2344994}. In the context of deep learning, \textcolor{Black}{RNN} and \textcolor{Black}{Long Short-term Memory (LSTM) Network} were proposed to solve the time-series prediction problem \cite{Hochreiter:1997:LSM:1246443.1246450}. \textcolor{Black}{Their }variants, such as Gated RNN \cite{DBLP:journals/corr/ChoMBB14}, were proposed in machine translation, which \textcolor{Black}{results in} similar performance compared with LSTM. \textcolor{Black}{Bidirectional LSTM (Bi-LSTM)} was proposed to capture both the forward and backward \textcolor{Black}{temporal properties} of the sequence \cite{Schuster:1997:BRN:2198065.2205129}. The attention mechanism was further incorporated into the Bi-LSTM in the paper \cite{Bahdanau2016} \textcolor{Black}{ to enable the model to assign} different weights to the historical data when predicting or translating. Temporal dependency plays a significant role in predicting maximum \textcolor{Black}{internal }stress. Both maximum \textcolor{Black}{internal }stress and  \textcolor{Black}{fracture propagation} have temporal features, \textcolor{Black}{which need to be fused and incorporated into designing the deep learning surrogates for prediction of maximum internal stress.} 

\textcolor{Black}{Apart from extracting features from historical data, the challenge of predicting maximum internal stress also lies in fusing spatial and temporal features. }There are recent advances in \textcolor{Black}{feature fusion in other fields}. Wang et al. \cite{NIPS2017_6689,DBLP:journals/corr/abs-1804-06300} proposed to combine CNN and LSTM networks to \textcolor{Black}{predict the entire video} based on \textcolor{Black}{the} initial few frames. Yao et al. \cite{DBLP:journals/corr/abs-1802-08714}, Wei et al. \cite{Wei:2018:IRL:3219819.3220096}, \textcolor{Black}{and} Zhang et al. \cite{Zhang:2019:CMR:3308558.3314139} proposed to use CNN to represent the spatial view of the city topology and \textcolor{Black}{to} use LSTM to represent the temporal view of traffic flow for predicting traffic condition. In \textcolor{Black}{predicting maximum internal stress,  historical stress data does not have sufficient information for future prediction, especially for multi-step prediction. Spatial and temporal properties of fracture propagation serve as necessary and} important \textcolor{Black}{ supplementary information} to reduce error accumulation in \textcolor{Black}{the} multi-step prediction. Incorporating the dynamic changes of \textcolor{Black}{fracture} into the prediction of maximum \textcolor{Black}{internal} stress is \textcolor{Black}{a} key challenge. 

\textcolor{Black}{This work proposes} a novel deep learning model, StressNet, to predict the maximum internal stress in the fracture propagation process. Instead of deterministically calculating  the entire stress field \textcolor{Black}{at each time step} as HOSS does, \textcolor{Black}{StressNet} focuses on predicting only the maximum internal stress, which \textcolor{Black}{is the key factor influencing} material failure. \textcolor{Black}{Spatial features of fractures, which are extracted by a Temporal Independent Convolutional Neural Network (TI-CNN), are incorporated} to help with the multi-step prediction of the maximum \textcolor{Black}{internal} stress. StressNet also \textcolor{Black}{uses} the Bi-directional LSTM (Bi-LSTM) \cite{Schuster:1997:BRN:2198065.2205129} to capture the temporal \textcolor{Black}{features} of \textcolor{Black}{fracture propagation and historical }maximum \textcolor{Black}{internal} stress. Finally, \textcolor{Black}{StressNet predicts} the \textcolor{Black}{future} maximum \textcolor{Black}{internal} stress by fusing the \textcolor{Black}{aforementioned spatial and temporal features.} During the training process, the Mean Squared Error (MSE) and Mean Absolute Percentage Error (MAPE) \textcolor{Black}{are integrated} as an adaptive objective function, with a dynamically tuned weight coefficient, to predict both the peak and bottom values \textcolor{Black}{better}. To the best of our knowledge, StressNet is the first work trying to use Deep Learning models to predict maximum internal stress with \textcolor{Black}{fracture propagation}. Inspired by physic knowledge and existing works in other domains, \textcolor{Black}{the StressNet} is designed to incorporate \textcolor{Black}{features from fracture propagation} into prediction \textcolor{Black}{and} fuse spatial and temporal features from multiple data formats.

\textcolor{Black}{T}his \textcolor{Black}{manuscript} is organized as follows: Section \ref{problem} introduces the details of HOSS simulations and the problem formulation. Section \ref{ML}  \textcolor{Black}{describes the proposed deep learning model,} StressNet. Section \ref{results} discusses \textcolor{Black}{the} experiment\textcolor{Black}{al} details and results. Section \ref{conclusion} summarizes th\textcolor{Black}{is work} and explores future directions.

\section{High-fidelity Simulations and Problem Formulation} \label{problem}

This section \textcolor{Black}{contains a discussion of} the high-fidelity simulations and problem formulation. 

\subsection{Hybrid Optimization Software Suite (HOSS) Simulations}

\textcolor{Black}{The data used for building and validating the ML model are from two-dimensional} HOSS simulations. Each simulation \textcolor{Black}{is conducted on a }rectangular sample material of $2m$ width  and $3m$ length, loaded with uniaxial tension. At the \textcolor{Black}{beginning of} each simulation, the material \textcolor{Black}{sample is seeded with} 20 cracks \textcolor{Black}{that
mimic} initial defects in the material. Each of the initial cracks shares the same length of 20 $cm$, and the orientation is chosen uniformly randomly to be 0, 60, or 120 degrees from horizontal. To keep the \textcolor{Black}{initial }cracks from overlapping, the \textcolor{Black}{material sample} is divided uniformly into 24 \textcolor{Black}{grids}, and 20 of \textcolor{Black}{them are randomly picked to place initial cracks}. As the simulation progresses, some initial cracks propagate and coalesce due to the external tensile loading. The material \textcolor{Black}{sample }completely fails when there is a single crack spanning the \textcolor{Black}{sample horizontally}. At this point, the material \textcolor{Black}{can not} carry any load.

\begin{figure*}[ht]
    \includegraphics[width=0.8\linewidth]{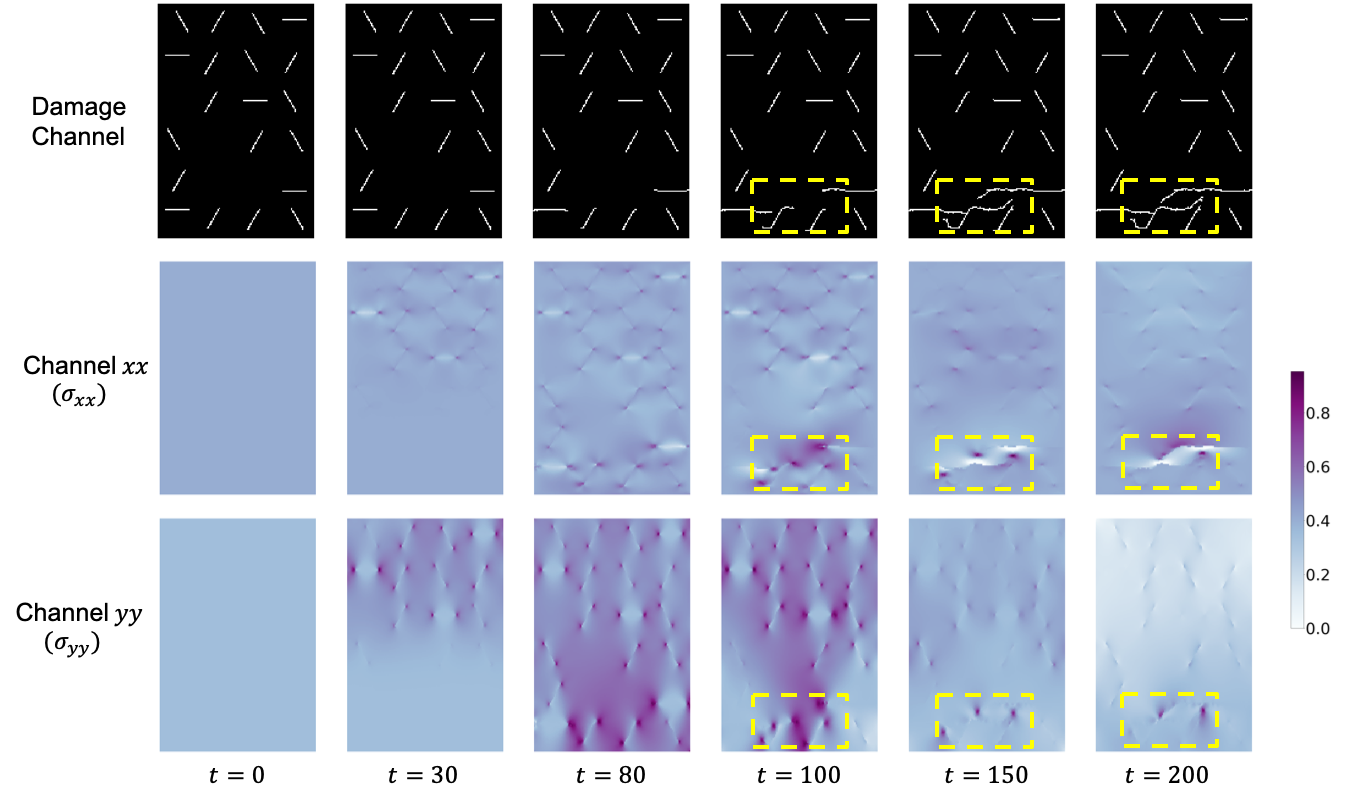}
    \centering 
    \caption{Visualization of the Fracture Growth and the Dynamic Changes of the Stress Field in Two Directions. The first row is the propagation of cracks and the simulation ends when a single crack span\textcolor{Black}{s} the width of the material, which is shown in the yellow-highlighted region in the first row. The white lines in the first row represent cracks and the black background represents the normal material. The second row represents the stress field of $\sigma_{xx}$ and the third row represents the stress field of $\sigma_{yy}$. According to the yellow dashed boxes highlighted in this figure, the stress tends to concentrate on the tips of existing cracks, and moreover, the cracks will grow because of the stress concentration.}
    \label{Crack_Sequence}
\end{figure*}

At each time step, the HOSS simulation outputs \textcolor{Black}{a 2-way tensor (matrix)} representing the position of current cracks and a 3-way tensor \textcolor{Black}{representing} the entire stress field. The \textcolor{Black}{sample data provided by the simulation} is shown in Fig. \ref{Crack_Sequence}. \textcolor{Black}{The first row of Fig. \ref{Crack_Sequence} is the distribution of cracks at each time step, and it is denoted as the damage channel. The second and third rows are the stress field,  decomposed into two directions, Channel $xx$ ($\sigma_{xx}$) and Channel $yy$ ($\sigma_{yy}$).} \textcolor{Black}{To help with easier visualization,} the stress field has been normalized into the range $[0,1]$ in both directions \textcolor{Black}{independently}. Also, the yellow dashed boxes in Fig. \ref{Crack_Sequence} show that stress tends to concentrate on the tips of cracks. Moreover, those cracks propagate when the maximum local stress exceeds a threshold. \textcolor{Black}{For more details about} HOSS\textcolor{Black}{, the reader can} refer to \cite{osti_1369045, SCHWARZER2019322, Lei2016}.

\subsection{Problem Formulation}
\label{Problem_Formulation}
\textcolor{Black}{Since} the high-fidelity HOSS model is \textcolor{Black}{computationally intensive} (each simulation takes about 4 hours on 400 processors), \textcolor{Black}{a deep learning model, StressNet, is proposed as a surrogate} to predict the maximum \textcolor{Black}{internal} stress \textcolor{Black}{until material} failure. Instead of the entire stress field, \textcolor{Black}{StressNet focuses} on the maximum \textcolor{Black}{internal} stress, which is highly correlated with \textcolor{Black}{fracture propagation.}
This is analogous to the relationship between the spring deformation and the external force. Consequently, in StressNet, the cracks' information \textcolor{Black}{is incorporated} to improve the accuracy in multi-step predictions. So that the input of the model consists of two parts, $x_{1}, ... , x_{\Delta t}$ denotes the $\Delta t$ consecutive time-steps of maximum \textcolor{Black}{internal} stress, and $I_{1}, ... , I_{\Delta t}$ denotes the \textcolor{Black}{fractures'} information in the same period. The \textcolor{Black}{criteria for determining the} value of $\Delta t$ \textcolor{Black}{is to} find the minimum input length containing sufficient temporal \textcolor{Black}{features} to make predictions. The \textcolor{Black}{cracks'} information is in matrix format at each time step, \textcolor{Black}{and it is} named as \textcolor{Black}{the} \textit{damage channel} \textcolor{Black}{in the rest of this paper}. The output of the model is \textcolor{Black}{the predicted internal} stress at the next time step, which is \textcolor{Black}{denoted as} $\hat{x}_{\Delta t + 1}$. To get the multi-step predictions towards the end of \textcolor{Black}{one }simulation (\textcolor{Black}{when} the material fails), the result from \textcolor{Black}{the} former step $\hat{x}_{\Delta t + 1}$ \textcolor{Black}{is fed} into the model to make further predictions. % Need to specify that the predictions will be unrolled to go beyond one step and give an idea of what size of \Delta t

\section{StressNet: A Deep Learning Model for the Prediction of Maximum Internal Stress} \label{ML}

StressNet is a \textcolor{Black}{novel} deep learning model that incorporates \textcolor{Black}{both the spatial and temporal features }of fracture propagation \textcolor{Black}{for maximum internal stress prediction}. Specifically, the Temporal Independent CNN (TI-CNN) captures spatial features of \textcolor{Black}{the damage channel} at each time step, and the Bidirectional LSTM (Bi-LSTM) \textcolor{Black}{is adapted} to capture the temporal features in the fracture \textcolor{Black}{propagation and historical stress data}. Spatial features in the stress field indicate the material fracture path \textcolor{Black}{and distributions} at each time step. Temporal features \textcolor{Black}{both describe} the \textcolor{Black}{dynamic properties of }fracture growth, which rely on external loadings and material \textcolor{Black}{physical} properties, \textcolor{Black}{and contain the features from historical stress data}. \textcolor{Black}{The proposed StressNet makes full use of various data formats (damage channel and stress field) of HOSS simulation outputs, and fuses the spatial and temporal features of those data. In this way}, the maximum \textcolor{Black}{internal }stress \textcolor{Black}{in the future} time-steps can be accurately predicted. \textcolor{Black}{In this section,} data properties \textcolor{Black}{are first summarized, which inspire the architecture of StressNet. Building blocks (TI-CNN and Bi-LSTM) of StressNet are then introduced. Finally, the detailed architecture of StressNet is discussed.}

\subsection{Data Properties}
\label{Data_Property}
\begin{itemize}
    \item \textbf{Significant Fluctuation:} The maximum \textcolor{Black}{internal} stress changes severely after the initial increase, as shown in red lines in Fig. \ref{Channel_xx_test_result} and \ref{Channel_yy_test_result}. There is no obvious trend of the changes in the forward direction. So the model needs to incorporate reference information, forward and backward temporal information to enrich features for the prediction \textcolor{Black}{of maximum internal stress.}
    
    \item \textbf{Spatial and Temporal Features:} The motivation for incorporating the damage channel into the maximum \textcolor{Black}{internal} stress prediction \textcolor{Black}{is introduced in section \ref{Problem_Formulation}}. The \textcolor{Black}{damage channels} within a certain time interval, $I_{1}, ..., I_{\Delta}$, contain both spatial and temporal features, \textcolor{Black}{and the historical stress data contains temporal features.} Our model \textcolor{Black}{is designed to capture} and fuse \textcolor{Black}{these} features.
    
    \item \textbf{Large Range:} The range of maximum stress data is from zero to \textcolor{Black}{a} scale of $10^{7}$. Even after \textcolor{Black}{normalization}, some of the data will be close to $0$, while some of the data will be close to $1$. Thus, our model needs to perform well on both \textcolor{Black}{peak and bottom} values to accurately \textcolor{Black}{predict} the stress change during the fracture propagation process.
\end{itemize}

\subsection{Temporal Independent Convolutional Neural Network (TI-CNN) on Damage Channel}
\label{TI-CNN}
\begin{figure}[ht]
    \centering 
    \includegraphics[width=0.7\linewidth]{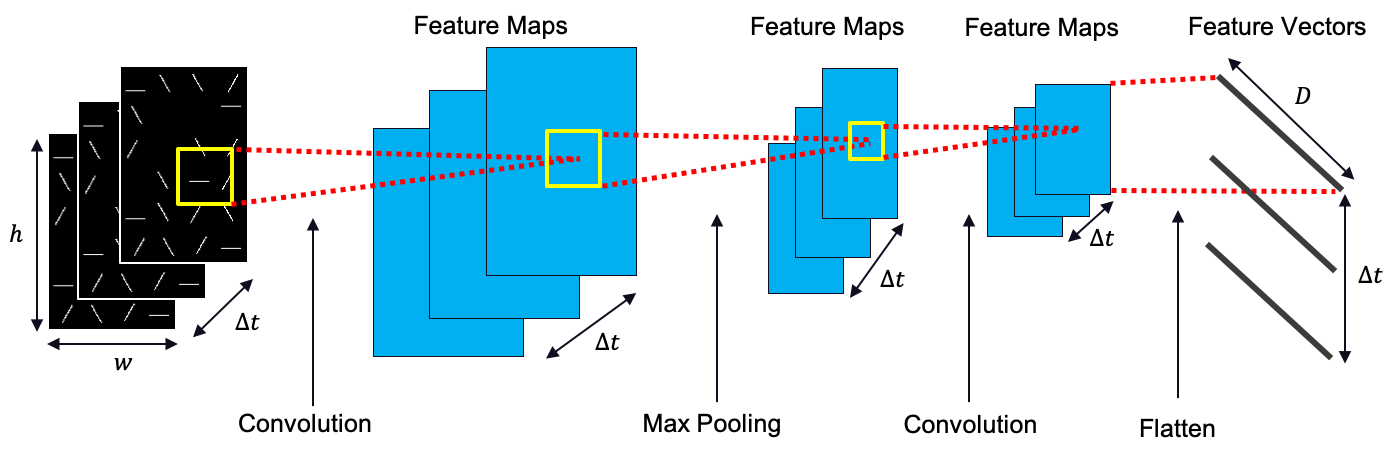}
    \caption{\textcolor{Black}{A}rchitecture of \textcolor{Black}{the} TI-CNN model. In TI-CNN, the independent convolution kernel for the damage channel at each time step \textcolor{Black}{is defined}. In this way, \textcolor{Black}{the} spatial features from the damage channel \textcolor{Black}{are extracted without changing} the temporal dependencies.}
    \label{CNN_Structure}
\end{figure}

Changes in the \textcolor{Black}{fracture pattern} could indicate changes in the maximum \textcolor{Black}{internal} stress; \textcolor{Black}{therefore}, the TI-CNN extracts spatial features \textcolor{Black}{from} the damage channel to \textcolor{Black}{represent} crack \textcolor{Black}{information} (including length, orientation, position, propagation, pairwise distance, etc.). As shown in Fig. \ref{CNN_Structure}, the input of TI-CNN is a 3-way tensor with a shape of $h\times w \times \Delta t$, \textcolor{Black}{in which the third dimension represents the time interval.} \textcolor{Black}{The original convolution operation is directly applied to the entire tensor, which will destroy the temporal dependencies \cite{NIPS2012_4824}. To keep temporal dependencies unchanged as well as extracting spatial features}, in TI-CNN, the independent convolution kernel \textcolor{Black}{is defined }for the damage channel at each time step. In a nutshell, the TI-CNN model consists of a stack of distinct layers, and the input tensor (damage channels \textcolor{Black}{within a certain interval) passes} through \textcolor{Black}{these} layers and outputs the features. Each of the layers \textcolor{Black}{is tailored for stress prediction }, which \textcolor{Black}{is} introduced below.

\subsubsection{Temporal Independent \textcolor{Black}{Convolutional} Layer}

Suppose that the input of the $l_{th}$ \textcolor{Black}{convolutional} layer is $\mathcal{X}_{l}$ with \textcolor{Black}{a} shape \textcolor{Black}{of} $H_{l} \times W_{l} \times \Delta t$, \textcolor{Black}{in which $H_{l}$ and $W_{l}$ represent the spatial dimension, and $\Delta t$ denotes the temporal dimension.} The corresponding convolutional kernel $\mathcal{K}_{l}$ has \textcolor{Black}{a} shape of $d \times d \times D_{l}$, \textcolor{Black}{in which $d$ represents the spatial size of the convolutional kernel, and $D_{l}$ denotes the number of kernels.} In the Temporal Independent Convolutional Layer, the number of kernels ($D_{l}$) \textcolor{Black}{is set }equal to the number of input channels ($\Delta t$), and the convolution operation \textcolor{Black}{is conducted} on each input channel separately. \textcolor{Black}{The expression of Temporal Independent Convolutional Layer is given as equation (\ref{3-2-1}).}

\textcolor{Black}{
\begin{align}
    \mathcal{X}_{l+1}(x,y,t) = \sum_{i = x}^{x + d - 1}\sum_{j = y}^{y + d - 1} \mathcal{K}_{l}(i-x,j-y,t)\mathcal{X}_{l}(i,j,t)
    \label{3-2-1}
\end{align}
}

In \textcolor{Black}{equation (\ref{3-2-1}), $\mathcal{X}_{l+1} \in \mathbb{R}^{(H_{l}-d+1)\times (W_{l}-d+1)\times \Delta t}$ represents the output of the $l_{th}$ convolutional layer, which is also the input of the next layer.} 

In \textcolor{Black}{maximum internal }stress prediction, the initial input of \textcolor{Black}{TI-CNN} is a time-series damage channel \textcolor{Black}{has a shape of}  $h\times w\times \Delta t$, as Fig. \ref{CNN_Structure} shows. To capture spatial features and keep temporal dependencies unchanged, \textcolor{Black}{the temporal independent convolution operation is conducted on the damage channels. The output is further fed into the following pooling layer.}

\subsubsection{Pooling Layer}

The pooling layer \textcolor{Black}{is, generally,} a non-linear downsampling function. For the sake of consistency, suppose that the input of the $l_{th}$ pooling layer is $\mathcal{X}_{l}$ with \textcolor{Black}{a} shape \textcolor{Black}{of} $H_{l} \times W_{l} \times \Delta t$, and the kernel size of the pooling layer is $d \times d$. The output of this layer $\mathcal{X}_{l+1}$ has the shape \textcolor{Black}{$\frac{H_{l}}{d} \times \frac{W_{l}}{d} \times \Delta t$}. In general, there are two kinds of pooling layers: average-pooling and max-pooling. For the pooling operation, at first, each channel of the input tensor \textcolor{Black}{is divided} into non-overlapping partitions which share the same \textcolor{Black}{spatial dimension ($d\times d$)} as the kernel. \textcolor{Black}{Then}, for the average pooling, the mean value of each partition \textcolor{Black}{is calculated}, while for the max-pooling, the maximum value of each partition \textcolor{Black}{is calculated}.

\textcolor{Black}{Fractures} only \textcolor{Black}{take up} a small area in the material; therefore, if average pooling \textcolor{Black}{is used}, then the features \textcolor{Black}{of} the large undamaged area would dilute or even hide the features \textcolor{Black}{of} the small damaged area. \textcolor{Black}{Therefore}, max-pooling \textcolor{Black}{is used} in \textcolor{Black}{ TI-CNN} to amplify the \textcolor{Black}{features of fracture }propagation.

\subsubsection{Fully Connected Layer}

The fully connected layer takes the feature map from the previous layer as the input and outputs the feature vector by matrix multiplication. Suppose that the input of the fully connected layer is $\mathcal{X}_{l}$ with \textcolor{Black}{a} shape \textcolor{Black}{of} $H_{l} \times W_{l} \times \Delta t$, \textcolor{Black}{it is at first reshaped into $\Tilde{X}_{l} \in \mathbb{R}^{H_{l}W_{l}\times \Delta t}$, and }the output of this layer is \textcolor{Black}{calculated in equation (\ref{3-2-2}).}

\textcolor{Black}{
\begin{align}
    X_{l+1} = W\Tilde{X}_{l}
    \label{3-2-2}
\end{align}
In equation (\ref{3-2-2}), $W \in \mathbb{R}^{D\times H_{l}W_{l}}$ represents the weight matrix in the fully connected layer, and $X_{l+1} \in \mathbb{R}^{D\times \Delta t}$ represents the output of the fully connected layer.}

The output of the \textcolor{Black}{TI-CNN} has \textcolor{Black}{a}  shape \textcolor{Black}{of} $D \times \Delta t$, which is the time series feature vector containing the spatial properties and preserving temporal dependencies of material fracture propagation.

\subsection{Bidirectional LSTM (Bi-LSTM) on Temporal Dependency}
\label{Bi-LSTM}
\textcolor{Black}{Capturing temporal dependencies is essential in predicting maximum internal stress with fracture propagation. } LSTM\textcolor{Black}{s} \cite{Hochreiter:1997:LSM:1246443.1246450} \textcolor{Black}{are} efficient variants of \textcolor{Black}{recurrent neural networks} which \textcolor{Black}{can} selectively remember \textcolor{Black}{the }immediate history of the input sequence \textcolor{Black}{and} longer-term trends. However, LSTMs only consider the forward pass over an input sequence; \textcolor{Black}{so} the prediction error accumulates when the former prediction results \textcolor{Black}{are used to make multi-step predictions. To reduce the accumulated error, each step prediction must be} as precise as \textcolor{Black}{possible}: not only consistent with the forward property (\textcolor{Black}{from past to future}) but also consistent with the backward property (\textcolor{Black}{from future to past}). Therefore, the temporal \textcolor{Black}{features in fracture propagation and historical stress data are captured} with a Bi-LSTM \cite{Schuster:1997:BRN:2198065.2205129}, \textcolor{Black}{to ensure that their predictions are consistent with forward and backward temporal dependencies}. 

\begin{figure}[ht]
    \includegraphics[width=0.7\linewidth]{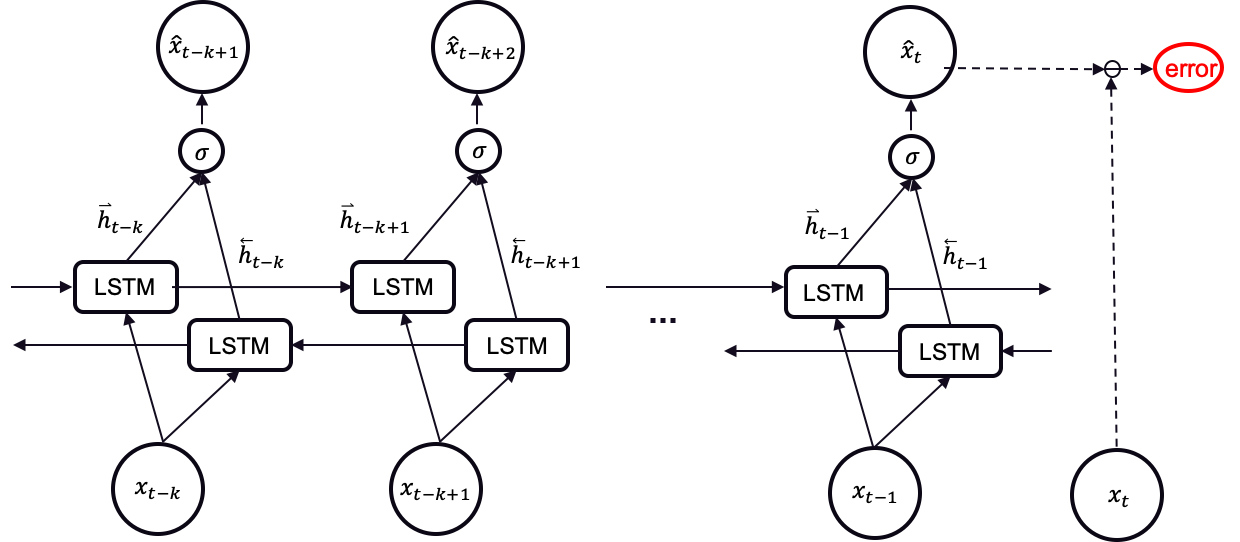}
    \centering
    \caption{Architecture of the Bi-LSTM}
    \label{BiLSTM_Structure}
\end{figure}

The structure of \textcolor{Black}{the} Bi-LSTM \cite{Schuster:1997:BRN:2198065.2205129} is shown in Fig. \ref{BiLSTM_Structure}, \textcolor{Black}{in which the model predicts the $\hat{x}_{t}$ given the input time series data $x_{t-k},...,x_{t-1}$.} Compared with LSTM, it has one extra hidden layer to capture the backward \textcolor{Black}{temporal properties within} the input data. \textcolor{Black}{More specifically, for maximum internal stress prediction, there are} two sources of time-series data \textcolor{Black}{serving as the input of Bi-LSTM}, one of them is the \textcolor{Black}{time-series }maximum \textcolor{Black}{internal }stress, and the other is the \textcolor{Black}{time-series }spatial features extracted from the damage channel \textcolor{Black}{(output of TI-CNN)}. The expressions of Bi-LSTM \textcolor{Black}{corresponding to Fig. \ref{BiLSTM_Structure} are} given below.

\begin{align}
    \overrightarrow{h}_{t} &= f(W_{1}x_{t} + W_{2}\overrightarrow{h}_{t-1} + b_{1})
    \notag\\
    \overleftarrow{h}_{t} &= f(W_{3}x_{t} + W_{4}\overleftarrow{h}_{t-1} + b_{2})
    \notag\\
    x_{t} &= \sigma(W_{5}\overrightarrow{h}_{t} + W_{6}\overleftarrow{h}_{t} + b_{3})
    \label{3-3-1}
\end{align}

\textcolor{Black}{In equations \ref{3-3-1}, the $\overrightarrow{h}_{t}$ and $\overleftarrow{h}_{t}$ represent the forward and backward temporal feature vectors at time $t$, respectively; $W_{i}, i=1,...,6$, denotes the weight matrices in the Bi-LSTM, $b_{j}, j=1,2,3,$ represent biases; \textcolor{Black}{and} $x_{t}$ represents the prediction result.}

\textcolor{Black}{In maximum internal stress prediction}, the stress \textcolor{Black}{data have} significant \textcolor{Black}{variations and do} not have an apparent trend in the first few time-steps, which makes the \textcolor{Black}{multi-step }prediction challenging. So Bi-LSTM \textcolor{Black}{is adapted} to capture the complex temporal dependency. In general, the Bi-LSTM is mainly used for two purposes. \textcolor{Black}{First}, separate Bi-LSTMs \textcolor{Black}{are adapted }to encode the historical maximum \textcolor{Black}{internal }stress \textcolor{Black}{and} the time-series \textcolor{Black}{spatial feature vectors (extracted from }damage channels), \textcolor{Black}{ and output} fixed-\textcolor{Black}{dimension time-series feature} vectors. \textcolor{Black}{Second}, \textcolor{Black}{the time-series feature vectors} from two data sources \textcolor{Black}{are fused and decoded to make }\textcolor{Black}{a} prediction. \textcolor{Black}{Detailed explanations} will be given in section \ref{Convolutional_Aided_BiLSTM}.

Based on \textcolor{Black}{the coupling effect between maximum internal stress and fracture propagation}, it is hard to predict \textcolor{Black}{the future maximum internal stress }purely based on previous stress data. Therefore, \textcolor{Black}{the time-series spatial feature vectors} extracted from the damage channel \textcolor{Black}{is incorporated }to improve the performance of prediction. \textcolor{Black}{In next section,} the structure of StressNet \textcolor{Black}{is introduced }to encode the dynamic properties of the maximum \textcolor{Black}{internal }stress and fuse the spatial \textcolor{Black}{internal and temporal features}.

\subsection{StressNet: Convolutional Aided Bidirectional LSTM}
\label{Convolutional_Aided_BiLSTM}
In sections \ref{TI-CNN} and \ref{Bi-LSTM}, the basic building blocks of \textcolor{Black}{StressNet are introduced}. TI-CNN is mainly used to capture the spatial features of \textcolor{Black}{the }damage channel, and Bi-LSTM is mainly used to encode the temporal dependencies of historical data. \textcolor{Black}{StressNet is proposed }to fuse the features from these building blocks \textcolor{Black}{and} improve the multi-step prediction performance.

\begin{figure*}[ht]
    \includegraphics[width=0.7\linewidth]{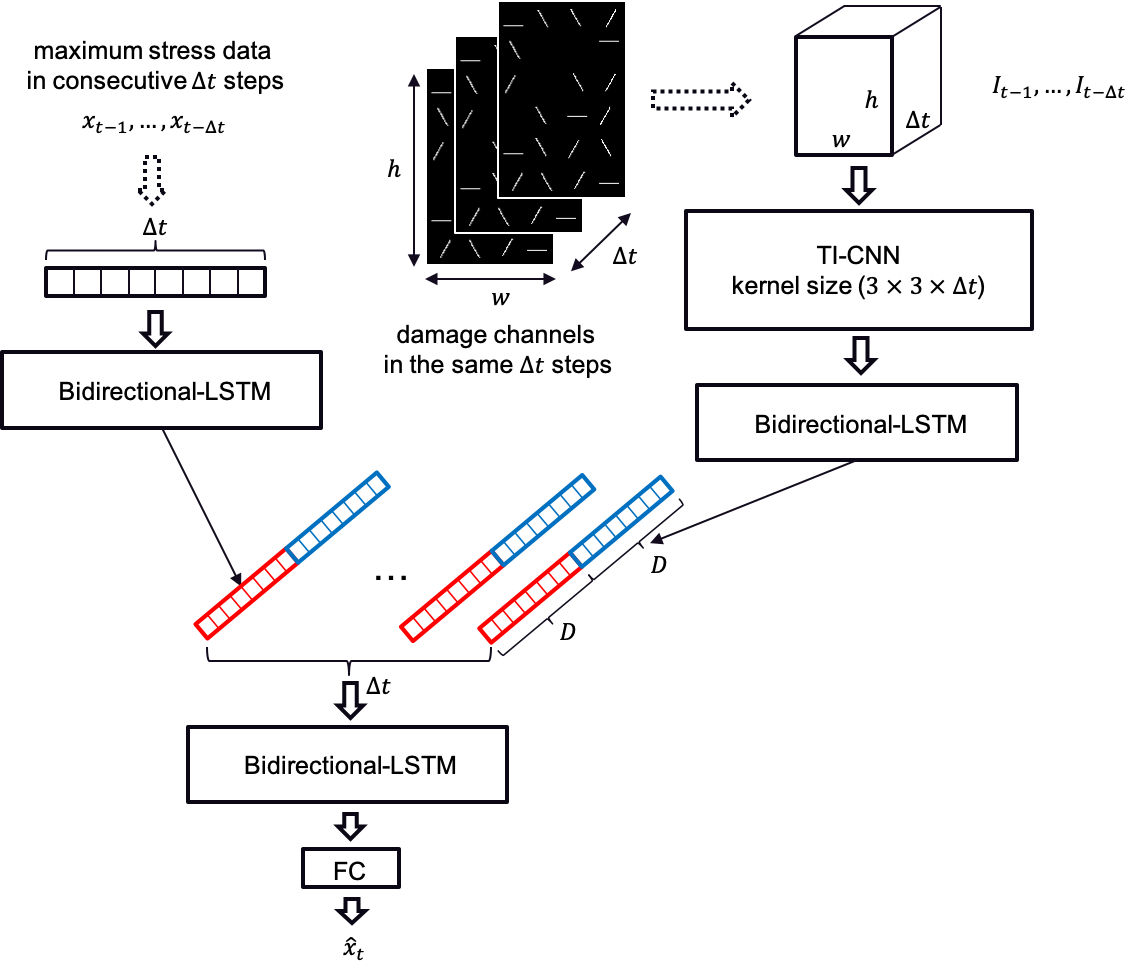}
    \centering
    \caption{Architecture of StressNet. The model consists of two branches. In the left branch, the Bi-LSTM encodes the temporal dependency among historical maximum stress data into a series of vectors. In the right branch, TI-CNN followed by Bi-LSTM encodes the spatial and temporal information of the damage channel into another series of vectors. By fusing features from these two branches, StressNet can predict the maximum stress at the next time step.}
    \label{Conv_Aided_BiLSTM}
\end{figure*}

\subsubsection{Model Structure}
The structure of StressNet is shown in Fig. \ref{Conv_Aided_BiLSTM}. \textcolor{Black}{The} goal is to predict the maximum \textcolor{Black}{internal }stress at the next time step, given the \textcolor{Black}{previous} maximum \textcolor{Black}{internal }stress and damage \textcolor{Black}{channels}, which \textcolor{Black}{is shown in equation} \eqref{4-3-1}. The left branch of \textcolor{Black}{StressNet} uses the Bi-LSTM to capture \textcolor{Black}{the bidirectional temporal properties }of historical maximum \textcolor{Black}{internal} stress. Suppose that initially, \textcolor{Black}{there are} consecutive $\Delta t$ steps of \textcolor{Black}{stress} data $x_{t-1}, x_{t-2}, ... , x_{t-\Delta t}$, the Bi-LSTM will encode their temporal dependencies into \textcolor{Black}{time-series} feature \textcolor{Black}{vectors} with a shape of \textcolor{Black}{$D\times \Delta t$} (as the red part shown in Fig. \ref{Conv_Aided_BiLSTM}). The right branch of \textcolor{Black}{StressNet uses} TI-CNN to \textcolor{Black}{extract spatial features of} the damage channels within the same consecutive time-steps $I_{t-1}, I_{t-2}, ... , I_{t-\Delta t}$, and then further \textcolor{Black}{extract} the temporal features into \textcolor{Black}{time-series }vectors with the same shape \textcolor{Black}{$D\times \Delta t$} (as the blue part shown in Fig. \ref{Conv_Aided_BiLSTM}). Up to now, at each time step, \textcolor{Black}{there are} two feature vectors with the same shape \textcolor{Black}{$D$} representing the \textcolor{Black}{features} from \textcolor{Black}{the} stress data and  \textcolor{Black}{the }damage channel, respectively. Every pair of feature vectors \textcolor{Black}{are concatenated and fed} into the \textcolor{Black}{last} Bi-LSTM layer to fuse and decode their temporal information. Finally, the \textcolor{Black}{predicted} maximum \textcolor{Black}{internal} stress $\hat{x}_{t}$ \textcolor{Black}{is given by the final fully connected layer}.

\begin{align}
    \hat{x}_{t} = f(x_{t-1}, ... , x_{t-\Delta t}, I_{t-1}, ... , I_{t-\Delta t})
    \label{4-3-1}
\end{align}

\textcolor{Black}{In summary, StressNet is designed to take the historical stress data (vector) and damage channels (3-way tensor) as the input to predict the maximum internal stress in the next time step. To generate the entire sequence of stress data recursively, the previously predicted results will be fed into the StressNet to make further predictions. As a surrogate model of HOSS simulation, StressNet will be trained and validated on the simulations generated from HOSS. After training, accurate prediction of maximum internal stress is beneficial to ensure the material reliability and further applied to evaluate its residual life.}
\subsubsection{Loss Function}

The loss function is used to evaluate the difference between the ground truth and model prediction. The unknown parameters in our model \textcolor{Black}{are estimated} by minimizing the loss function. \textcolor{Black}{StressNet is tested} on three different loss functions, which are Mean Absolute Percentage Error (MAPE), Mean Squared Error (MSE), and dynamic fusion of MAPE and MSE.
\\

\textbf{(1) Mean Absolute Percentage Error (MAPE):} 

The expression of MAPE is given as equation \eqref{4-4-1}. In practice, the advantage of MAPE is that it is a relative loss which treats the large and small values equally. However, it is hard to get the minimum by using gradient descent methods because of the absolute component.

    \begin{align}
        \text{MAPE} = \frac{1}{T} \sum_{t = 1}^{T} \frac{|\hat{x}_{t} - x_{t}|}{x_{t}}
        \label{4-4-1}
    \end{align}
    
\textbf{(2) Mean Squared Error (MSE):}

Similarly, the MSE is \textcolor{Black}{calculated} as equation \eqref{4-4-2}. Unlike MAPE, the MSE will tend to perform better \textcolor{Black}{in} predicting large values \textcolor{Black}{while ignoring} small values to some extent. Also, the MSE is easy to optimize.

\begin{align}
    \text{MSE} = \frac{1}{T} \sum_{t = 1}^{T} (\hat{x}_{t} - x_{t})^{2}
    \label{4-4-2}
\end{align}

\textbf{(3) Dynamic Fusion of MAPE and MSE:}

In \textcolor{Black}{the} problem \textcolor{Black}{of maximum internal stress prediction}, the stress data fluctuate significantly \textcolor{Black}{with time}. \textcolor{Black}{Capturing} those fluctuations \textcolor{Black}{requires StressNet} to predict both the large and small values precisely. Moreover, \textcolor{Black}{StressNet should pay} more attention \textcolor{Black}{to} large values because \textcolor{Black}{they will have a major impact on} the \textcolor{Black}{fracture propagation}. Furthermore, \textcolor{Black}{the }loss function \textcolor{Black}{should} adapt to the stress fluctuation and be easy to converge. Based on these requirements, \textcolor{Black}{an adaptive loss function is designed as the} dynamic fusion of MAPE and MSE. The expression is given below.

\begin{align}
    \mathcal{L}(\theta, \beta) = \frac{1}{T} \sum_{t = 1}^{T} \left(\lambda(\beta)(\hat{x}_{t} - x_{t})^{2} + (1-\lambda(\beta))\frac{(\hat{x}_{t} - x_{t})^{2}}{x_{t}^{2}}\right)
    \label{4-4-3}
\end{align}

In equation \eqref{4-4-3}, $\theta$ represents the trainable parameters in \textcolor{Black}{StressNet}; $\beta$ represents the index of the current \textcolor{Black}{training} epoch. $\lambda$ is the hyper-parameter. It is the function of $\beta$ and is used to fuse the two components. The value of $\lambda$ \textcolor{Black}{is updated} during the training process. \textcolor{Black}{In general}, MSE tends to give better prediction on large values, while MAPE tends to perform better on small values. $\lambda$ \textcolor{Black}{is set to a large value} at the beginning of \textcolor{Black}{the} training process to get better performance on large target values. As the training process goes by, the value of $\lambda$ \textcolor{Black}{is decreased for improving prediction on the small target values}. This loss function is easy to optimize and \textcolor{Black}{is robust} to large and small values.

\section{Experiments and Results} \label{results}
In this section, the performance of the proposed StressNet \textcolor{Black}{is shown }by comparing it with other benchmark methods.

\subsection{Data Description and Preprocessing}

\textcolor{Black}{The dataset is composed of} 61 high-fidelity HOSS simulations, and each of them contains 228 \textcolor{Black}{time-}steps to simulate the detailed \textcolor{Black}{fracture propagation process}. Each simulation \textcolor{Black}{contains} the binary image data (damage channel) denoting the position of cracks at each time step, \textcolor{Black}{in} which 0 \textcolor{Black}{represents} undamaged material and 1 \textcolor{Black}{represents} damaged material. \textcolor{Black}{Each simulation also contains} the time-series maximum \textcolor{Black}{internal} stress. 

The original damage channel has a shape of $192 \times 128$, which makes the TI-CNN model large and \textcolor{Black}{challenging} to train. \textcolor{Black}{During the data preprocessing, }the damage channel \textcolor{Black}{is downsampled} into the shape $24 \times 16$ using the max-pooling method with filter size $8 \times 8$. The downsampled data can preserve the properties of cracks such as orientation, position, and dynamic changes of the crack length.

The time-series maximum \textcolor{Black}{internal }stress data\textcolor{Black}{set} has a \textcolor{Black}{wide} range, and the difference between the peak and bottom value could be up to $10^{7}$. To tackle this problem, the original data \textcolor{Black}{is normalized} into the range $[0,1]$ by using the min-max normalization method, which is given in equation \eqref{5-1-1}.

\begin{align}
    x_{t}^{\text{norm}} = \frac{x_{t} - x_{\text{min}}}{x_{\text{max}} - x_{\text{min}}},
    \label{5-1-1}
\end{align}
where $x_{\text{max}}$ and $x_{\text{min}}$ are the maximum and minimum stress data among all the simulations. \textcolor{Black}{The model  is trained and tested} by using the normalized data, and then \textcolor{Black}{the predicted results are reversed back} into the original value.

\textcolor{Black}{Furthermore}, in \textcolor{Black}{HOSS} simulations, the stress data at each time step is decomposed into three components, which are denoted as Channel $xx$, Channel $xy$, and Channel $yy$. Among all of them, Channel $xx$ and Channel $yy$ represent two stress components with orthogonal directions and determine the fracture propagation. In \textcolor{Black}{the experiment}, the same model structure \textcolor{Black}{is applied }to  predict Channel $xx$ and Channel $yy$ separately.

\subsection{Training Settings}

\textcolor{Black}{The code is implemented using the Python libraries} Keras \cite{chollet2015keras} \textcolor{Black}{and} Tensorflow \cite{tensorflow2015-whitepaper}. During the training phase, \textcolor{Black}{the adaptive moment estimator known as the Adam optimizer \cite{Kingma2014AdamAM} is used}, and the learning rate \textcolor{Black}{is set} at $10^{-3}$. \textcolor{Black}{Note that Adam is an optimizer based on gradient and momentum, which is used to minimize the loss function and update the weight matrices in StressNet. The dataset contains } 61 \textcolor{Black}{groups }of high-fidelity HOSS simulations, \textcolor{Black}{and }55 of them \textcolor{Black}{are selected as} the training data \textcolor{Black}{to build the model}. \textcolor{Black}{The remaining simulations are used to test model performance after training. In the training process, one epoch represents that the model was trained once throughout the entire training dataset. During each epoch, the dataset is ordered randomly and split into batches. Generally, the training process contains multiple epochs. At each epoch, six simulations are randomly selected and set aside for model validation. Note that validation is conducted at the end of each epoch to assess the model's performance. The \textcolor{Black}{goal of} validation is to indicate the model performance in data unseen during the epoch to avoid overfitting. This is different from testing --- which is conducted just once after all training --- on data never fed to the network during the training process}. 

\textcolor{Black}{In summary}, at each epoch, \textcolor{Black}{StressNet is trained} on 49 simulations, \textcolor{Black}{and validated on six simulations. After training, the model is tested} on the \textcolor{Black}{remaining} six simulations. \textcolor{Black}{To} prevent overfitting, the order of \textcolor{Black}{feeding} simulations \textcolor{Black}{is shuffled} every 30 epochs, \textcolor{Black}{and the shuffling process is repeated} 60 times, which means \textcolor{Black}{there are  1,800 epochs total}. When \textcolor{Black}{the} dynamic loss function \textcolor{Black}{is applied}, the value of $\lambda$ \textcolor{Black}{is set to} $0.9$ \textcolor{Black}{for the first 600 epochs}, and \textcolor{Black}{then, it is changed to} $0.1$ \textcolor{Black}{for the remaining simulations}. The training process takes between 8 to 20 hours on a single \textcolor{Black}{NVIDIA GeForce GTX 1080Ti GPU, depending on} the number of epochs.

\textcolor{Black}{The training phase is conducted on} one-step prediction, which means that \textcolor{Black}{StressNet only} predicts one step, and \textcolor{Black}{then it} compares the result with the ground truth. The input and output of the model at \textcolor{Black}{the} training phase \textcolor{Black}{are} shown in Table \ref{table.5-2-1}, in which $I_{t}$ is the damage channel at time $t$, $x_{t}^{\text{norm}}$ is the ground truth at time $t$, and $\hat{x}_{t}^{\text{norm}}$ denotes the prediction at time $t$. In \textcolor{Black}{the simulations}, the number of time-steps is $T = 228$, and $\Delta t = 10$. The validation phase has the same setting as the training phase.

\begin{table}[ht]
    \centering
    \caption{Input and Output at the Training Phase. \textcolor{Black}{T}he true maximum stress data and downsampled damage channel \textcolor{Black}{are used} as the \textcolor{Black}{model's} input, and \textcolor{Black}{the model's} output \textcolor{Black}{is compared} with \textcolor{Black}{the} ground truth.}
    \begin{tabular}{cc}
    \hline
    Input& Output\\
    \hline\\
    $(x_{1}^{\text{norm}}, I_{1}), (x_{2}^{\text{norm}}, I_{2}), ... , (x_{\Delta t}^{\text{norm}}, I_{\Delta t})$ & $\hat{x}_{\Delta t + 1}^{\text{norm}}$ \\\\
    $(x_{2}^{\text{norm}},I_{2}), (x_{3}^{\text{norm}}, I_{3}), ... , (x_{\Delta t+1}^{\text{norm}}, I_{\Delta t+1})$ & $\hat{x}_{\Delta t + 2}^{\text{norm}}$ \\\\
    \dots & \dots \\\\
    $(x_{T - \Delta t - 1}^{\text{norm}}, I_{T - \Delta t - 1}), (x_{T - \Delta t}^{\text{norm}},I_{T - \Delta t}), ... , (x_{T - 1}^{\text{norm}},I_{T - 1})$ & $\hat{x}_{T}^{\text{norm}}$ \\\\
    \hline
    \end{tabular}
    \label{table.5-2-1}
\end{table}

\subsection{Testing}

The training phase \textcolor{Black}{can be conducted on the} one-step prediction because \textcolor{Black}{the entire time-series \textcolor{Black}{stress data} are provided to }train the model. However, \textcolor{Black}{in} the testing phase, only the initial $\Delta t$ steps of stress data\textcolor{Black}{, $x_{1}^{\text{norm}},...,x_{\Delta t}^{\text{norm}}$, }\textcolor{Black}{are available. Hence, the model has} to make predictions recursively by successively using \textcolor{Black}{the} former prediction \textcolor{Black}{ $\hat{x}_{t}^{\text{norm}}$} to predict \textcolor{Black}{the maximum internal stress $\hat{x}_{t+1}^{\text{norm}}$ in }the next time-step. The input and output at the testing phase are shown in Table \ref{table.5-2-2}\textcolor{Black}{, in which, start from the second row, the previous predictions are served as the model input.} It take\textcolor{Black}{s approximately} 20 seconds to generate one entire simulation (228 time-steps).

\begin{table}[ht]
    \centering
    \caption{Input and Output at the Testing Phase. In the testing phase, only the initial $\Delta t$ steps of true maximum stress data \textcolor{Black}{are fed into the model, and} the true downsampled damage channel \textcolor{Black}{is used}. And then, \textcolor{Black}{the} most recent prediction \textcolor{Black}{is fed} into the model. \textcolor{Black}{This process is repeated until the} entire \textcolor{Black}{time-series is generated}.}
    \begin{tabular}{cc}
    \hline
    Input& Output\\
    \hline\\
    $(x_{1}^{\text{norm}}, I_{1}), (x_{2}^{\text{norm}}, I_{2}), ... , (x_{\Delta t}^{\text{norm}}, I_{\Delta t})$ & $\hat{x}_{\Delta t + 1}^{\text{norm}}$ \\\\
    $(x_{2}^{\text{norm}},I_{2}), (x_{3}^{\text{norm}}, I_{3}), ... , (\hat{x}_{\Delta t + 1}^{\text{norm}}, I_{\Delta t+1})$ & $\hat{x}_{\Delta t + 2}^{\text{norm}}$ \\\\
    \dots & \dots \\\\
    $(\hat{x}_{T - \Delta t - 1}^{\text{norm}}, I_{T - \Delta t - 1}), (\hat{x}_{T - \Delta t}^{\text{norm}},I_{T - \Delta t}), ... , (\hat{x}_{T - 1}^{\text{norm}},I_{T - 1})$ & $\hat{x}_{T}^{\text{norm}}$ \\\\
    \hline
    \end{tabular}
    \label{table.5-2-2}
\end{table}

\subsection{Baseline Models}
StressNet has two characteristics. One is \textcolor{Black}{that the} damage channel \textcolor{Black}{is incorporated} as reference information to \textcolor{Black}{improve accuracy in} multi-step predictions on maximum \textcolor{Black}{internal} stress. The other is \textcolor{Black}{that} the MAPE and MSE \textcolor{Black}{are adaptively fused as loss functions} according to the data properties. \textcolor{Black}{To show the performance of the proposed StressNet, we selected the diverse baseline models as benchmark. To ensure a fair comparison, all the benchmark methods are trained using the same Adam optimizer.}

\begin{itemize}
  \item \textbf{Historical Average:} Historical average predicts the maximum \textcolor{Black}{internal} stress at time step $t$ by using the average value of all simulations at the same time step. In \textcolor{Black}{the} experiments, the average of all training simulations at each time step \textcolor{Black}{is calculated} and \textcolor{Black}{used} as the prediction of the test data.
  \item \textbf{LSTM:} LSTM \cite{Hochreiter:1997:LSM:1246443.1246450} is a popular method for time series prediction, which combines the long term and short term temporal dependencies to make the prediction. In \textcolor{Black}{the} experiment, it is hard for the LSTM to give a reasonable prediction of each simulation (228 time-steps in all) only based on the initial \textcolor{Black}{ten time-steps} of data. \textcolor{Black}{So }in the \textcolor{Black}{LSTM, the value of }$\Delta t$ \textcolor{Black}{is set to 50}, which means that the initial 50 \textcolor{Black}{time-steps} of data \textcolor{Black}{are used }as the input \textcolor{Black}{to recursively predict the entire time-series}.
  \item \textbf{Bi-LSTM:} The structure of Bi-LSTM \cite{Schuster:1997:BRN:2198065.2205129} is shown in Fig. \ref{BiLSTM_Structure}, which takes both the forward and backward \textcolor{Black}{temporal properties} into consideration. Similar to the LSTM, we also set $\Delta t = 50$ for the Bi-LSTM.
  \item \textbf{StressNet + MSE:} \textcolor{Black}{The MSE} \textcolor{Black}{is used }as the loss function and keep the model structure the same as shown in Fig. \ref{Conv_Aided_BiLSTM}, which aims to compare the performance of different loss functions. 
  \item \textbf{StressNet + MAPE:} Another variant of the \textcolor{Black}{StressNet is} using MAPE as the loss function.
\end{itemize}

\subsection{Evaluation Metrics}

\textcolor{Black}{To evaluate the performances on predicting the peak and bottom values of maximum internal stress equally, the} MAPE \textcolor{Black}{is selected} to evaluate the performance of \textcolor{Black}{StressNet}. The expression of MAPE is given as equation \eqref{4-4-1}. \textcolor{Black}{According to section \ref{Data_Property}}, one of the important features of maximum \textcolor{Black}{internal} stress is that it fluctuates significantly. \textcolor{Black}{The MAPE is selected} to treat data with large and small value equally when \textcolor{Black}{evaluating} the model performance.

\subsection{Results and Discussion}

The numerical results of \textcolor{Black}{ StressNet} and baseline models are shown in Table. \ref{table.5-6-1}. \textcolor{Black}{Compared with the baseline models, }the proposed StressNet incorporates features from \textcolor{Black}{the} damage channel \textcolor{Black}{and uses the adaptively fusing} loss function. To show the benefit of incorporating the damage channel in multi-step predictions, several classical time series prediction models \textcolor{Black}{are selected,} including Historical Average, LSTM \cite{Hochreiter:1997:LSM:1246443.1246450}, and Bi-LSTM \cite{Schuster:1997:BRN:2198065.2205129}. \textcolor{Black}{The results show} that StressNet significantly outperforms these time series models even though the LSTM and Bi-LSTM \textcolor{Black}{took advantage of using more initial data for model training}.

In order to demonstrate the strength of the \textcolor{Black}{adaptively fusing} loss function, \textcolor{Black}{it is compared} with its two components MSE and MAPE. From the theoretical analysis, MSE performs better on large values, while MAPE performs better on small values. Among all the variants of the loss function, \textcolor{Black}{the results show that }the fused loss \textcolor{Black}{function achieves} the best performance, \textcolor{Black}{with an error of 2\%.} 

\begin{table}[ht]
    \centering
    \caption{Performances Comparison  Among Different Models by Using MAPE. All the variants of StressNet outperform other methods, and StressNet combined with Dynamic Fusion Loss Function received the best MAPE $\approx$ $2\%$.}
    \begin{tabular}{ccc}
    \hline
    Models& Channel $xx$ ($\sigma_{xx}$) & Channel $yy$ ($\sigma_{yy}$)\\
    \hline
    Historical Average & 0.0808& 0.0632\\
    LSTM & 0.1367 & 0.1023\\ 
    Bi-LSTM & 0.1103 & 0.0507\\
    StressNet(MSE) &0.0386 &0.0336\\
    StressNet(MAPE) &0.0394 &0.0340\\
    StressNet(Dynamic Loss) &\textbf{0.0218} &\textbf{0.0193}\\
    \hline
    \end{tabular}
    \label{table.5-6-1}
\end{table}

\begin{figure*}[!ht]
    \includegraphics[width=0.95\linewidth]{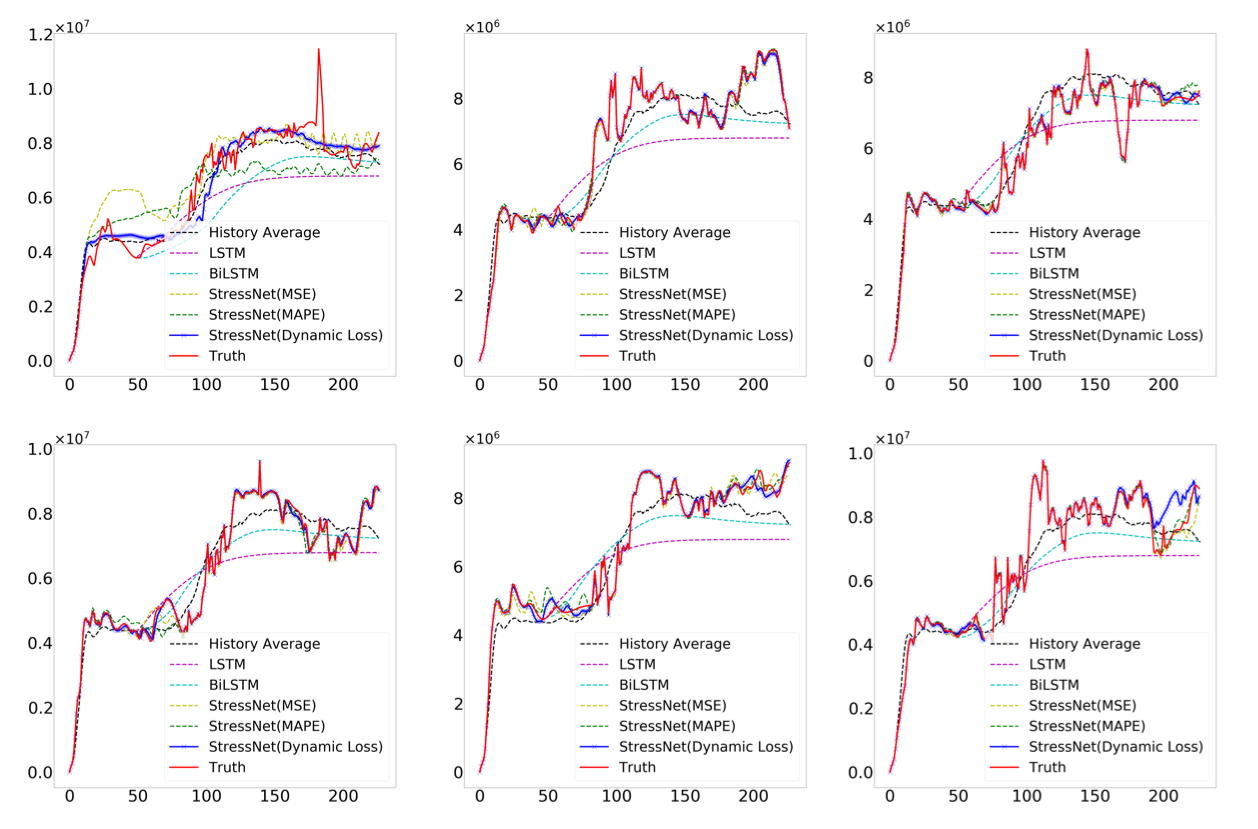}
    \centering
    \caption{Comparison of Test Results on Channel $xx$ ($\sigma_{xx}$) From Different Models. We use the same training and testing data to train and test each model. From the result, StressNet combined with Dynamic Fusion Loss Function receives the best performance (solid blue line).}
    \label{Channel_xx_test_result}
\end{figure*}

\begin{figure*}[!ht]
    \includegraphics[width=0.95\linewidth]{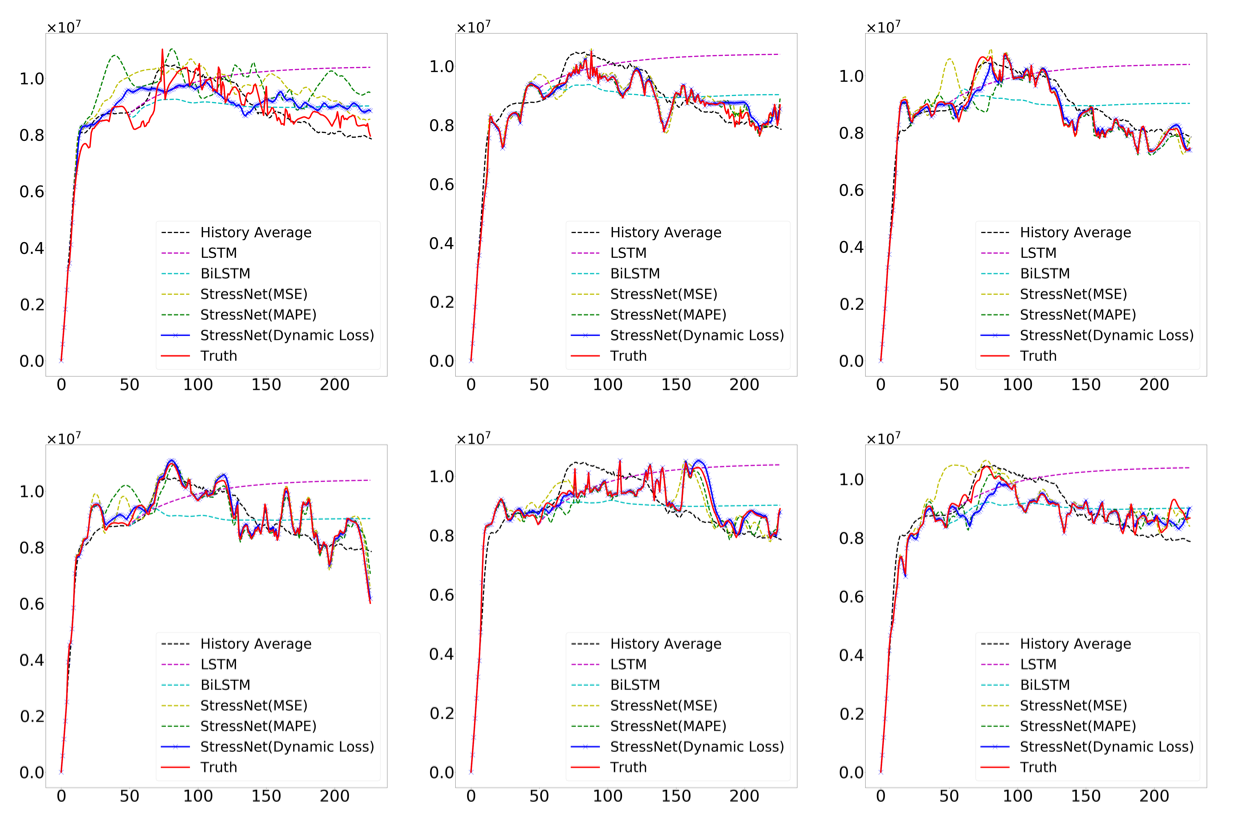}
    \centering
    \caption{Comparison of Test Results on Channel $yy$ ($\sigma_{yy}$) From Different Models. StressNet combined with Dynamic Fusion Loss Function has the best performance (solid blue line).}
    \label{Channel_yy_test_result}
\end{figure*}

\textcolor{Black}{In order to better visualize the experiment results}, the predictions of the different models on the test data and the corresponding ground truth \textcolor{Black}{are plotted} in Fig. \ref{Channel_xx_test_result} and Fig. \ref{Channel_yy_test_result} \textcolor{Black}{for channel $xx$ and channel $yy$},  respectively. \textcolor{Black}{The figures show }that the  ground truth plot \textcolor{Black}{shows severe fluctuations and orders of magnitude variation}. Although the normalization \textcolor{Black}{technique} makes it smoother, such properties \textcolor{Black}{are still} a challenge for \textcolor{Black}{multi-step} predictions. The results of the Historical Average \textcolor{Black}{show} that it successfully predicts the overall trend of changes but \textcolor{Black}{fails} to capture the \textcolor{Black}{fluctuations} and peaks, which are \textcolor{Black}{critical factors causing} fracture growth. Bi-LSTM \cite{Schuster:1997:BRN:2198065.2205129} and LSTM \cite{Hochreiter:1997:LSM:1246443.1246450} share a similar performance \textcolor{Black}{to} the Historical Average. It mainly because \textcolor{Black}{they} both tend to \textcolor{Black}{pay} more attention to the most recent data when making the prediction. If multi-step prediction \textcolor{Black}{is conducted} recursively, their outputs tend to converge to the same value. StressNet combined with \textcolor{Black}{the adaptively fusing} loss function \textcolor{Black}{successfully captures most of the fluctuations and receives accurate predictions on both peak and bottom values}. 

\section{Conclusion} \label{conclusion}
Accurate predictions of \textcolor{Black}{maximum} internal stress in brittle materials under dynamic loading \textcolor{Black}{conditions is critical for} advancing the fundamental understanding of the microstructure-sensitive fracture propagation mechanism \textcolor{Black}{and} ensuring system safety. Maximum internal stress is highly correlated with \textcolor{Black}{fracture propagation} within the material, and in turn, the presence of \textcolor{Black}{fractures} plays a key role in determining the level of internal stresses. Due to these complex nonlinear inter-dependencies, accurately predicting the maximum internal stress remains a challenging problem in material sciences.

 The \textcolor{Black}{novelty} of \textcolor{Black}{this} work lies in designing a physics-based model, \textcolor{Black}{StressNet}, which \textcolor{Black}{combines} the material fracture characteristics and the data properties within a deep learning architecture.  \textcolor{Black}{StressNet is designed to predict the entire sequence of maximum} \textcolor{Black}{internal} stress \textcolor{Black}{until material failure}. Unlike statistical learning method\textcolor{Black}{s} or those using manually selected features, \textcolor{Black}{ StressNet}  \textcolor{Black}{that} directly \textcolor{Black}{integrates} the damage channel into the multi-step predictions, and \textcolor{Black}{learns} the features \textcolor{Black}{by minimizing a} loss function. The advantages of \textcolor{Black}{the StressNet} could be summarized as follows. \textbf{(i)} After training, the model can generate the entire time series of maximum internal stress in \textcolor{Black}{about 20 seconds}, which significantly reduces the computation time from \textcolor{Black}{more than 4 hours} to 20 seconds  \textcolor{Black}{as compared to}  the high-fidelity HOSS model. \textbf{(ii)} Compared with statistical learning models, \textcolor{Black}{the} proposed model receives the best prediction performance with an error of 2\%. \textbf{(iii)} As a physics informed data-driven model,  \textcolor{Black}{StressNet} is flexible enough such that it is easy to generalize to other fracture propagation scenarios that involve diverse loading \textcolor{Black}{conditions} and different material  \textcolor{Black}{properties}.

\ifCLASSOPTIONcompsoc
  \section*{Author Contributions}
\else
  \section*{Author Contributions}
\fi

Yinan Wang mainly conducted the data preprocessing, model designing, and results comparison under the guidance of Diane Oyen and Xiaowei Yue. Weihong (Grace) Guo mainly offered ideas in results comparison. Anishi Mehta, and Cory Braker Scott offered the experimental data. Nishant Panda, M. Giselle Fernández-Godino, and Gowri Srinivasan offered the physical interpretation of data, provided ideas in model designing. All authors discussed the results and contributed to the final version of manuscript.

\ifCLASSOPTIONcompsoc
  \section*{Competing Interests}
\else
  \section*{Competing Interests}
\fi

The Authors declare no Competing Financial or Non-Financial Interests

\ifCLASSOPTIONcompsoc
  \section*{Code Availability}
\else
  \section*{Code Availability}
\fi

The "StressNet" code will be available upon request to the corresponding author. We are preparing the standardized package and will make it public later. 

\ifCLASSOPTIONcompsoc
  \section*{Data Availability}
\else
  \section*{Data Availability}
\fi

The data that support the findings of this study was generated in Los Alamos National Laboratory. Restrictions may apply to the availability of these data, Data are available from the authors upon
reasonable request and with permission of Los Alamos National Laboratory.

\ifCLASSOPTIONcompsoc
  \section*{Acknowledgments}
\else
  \section*{Acknowledgment}
\fi

Research supported by the Laboratory Directed Research and Development program of Los Alamos National Laboratory under project number 20170103DR and the LANL Applied Machine Learning Summer Research Fellowship.


\bibliographystyle{IEEEtran}
\bibliography{IEEEabrv,references}
\end{document}